\documentclass[10pt,twocolumn,letterpaper]{article}

\usepackage{cvpr}
\usepackage{times}
\usepackage{epsfig}
\usepackage{graphicx}
\usepackage{amsmath}
\usepackage{amssymb}
\usepackage{comment}
\usepackage{makecell}
\usepackage{enumitem}
\usepackage{comment}
\usepackage{tikz}
\usetikzlibrary{shapes.geometric}

\usepackage{multirow}

\usepackage[pagebackref=true,breaklinks=true,letterpaper=true,colorlinks,bookmarks=false]{hyperref}

 \cvprfinalcopy 

\def\cvprPaperID{1387} 

\ifcvprfinal\fi
\begin{document}

\title{Exploiting Image-trained CNN Architectures\\for Unconstrained Video Classification}

\author{Shengxin Zha\\
Northwestern University\\
Evanston IL USA\\
{\tt\small szha@u.northwestern.edu}
\and
Florian Luisier,
Walter  Andrews\\
Raytheon BBN Technologies\\
Cambridge, MA USA\\
{\tt\small \{fluisier,wandrews\}@bbn.com}\\
\and
Nitish Srivastava,
Ruslan Salakhutdinov\\
University of Toronto\\
{\tt\small \{nitish,rsalakhu\}@cs.toronto.edu}
}
\maketitle

\begin{abstract}
We conduct an in-depth exploration of different strategies for doing event detection
in videos using convolutional neural networks (CNNs) trained for
image classification. We study different ways of performing 
spatial and temporal pooling, feature normalization, choice of CNN layers
as well as choice of classifiers. Making judicious choices along these
dimensions led to a very significant increase in performance over more naive
approaches that have been used till now.  
We evaluate our approach on the challenging TRECVID MED'14 dataset
with two popular CNN architectures pretrained on ImageNet.
On this MED'14 dataset, our methods, based entirely on image-trained CNN
features, can outperform several state-of-the-art non-CNN models. 
Our proposed late fusion of CNN- and motion-based features 
can further increase the mean average precision (mAP) on MED'14 from $34.95\%$ to $38.74\%$.
The fusion approach achieves the state-of-the-art classification performance on the challenging UCF-101 dataset. 
\end{abstract}

\section{Introduction}
\label{sec:intro}
The huge volume of videos that are nowadays routinely produced by consumer cameras and shared across the web calls for effective video classification and retrieval approaches. 
One straightforward approach considers a video as a set of images and relies on techniques designed for image classification. 
This standard image classification pipeline consists of three processing steps: 
first, extracting multiple carefully engineered local feature descriptors (e.g., SIFT~\cite{lowe2004} or SURF~\cite{bay2006});
second, the local feature descriptors are encoded using the bag-of-words (BoW)~\cite{csurka2004} or Fisher vector (FV)~\cite{perronnin2010} representation;
and finally, classifier is trained (e.g., support vector machines (SVMs)) on top of the encoded features.
The main limitation of directly employing the standard image classification approach to video is the lack of exploitation of motion information.
This shortcoming has been addressed by extracting optical flow based descriptors (eg. \cite{lin2009opfl}), 
descriptors from spatiotemporal interest points ( e.g., \cite{laptev2005,dalal2005,willems2008,laptev2008,wang2009}) 
or along estimated motion trajectories 
(e.g., \cite{wang2011,jiang2012,jain2013,wang2013,wang2013action}).    

The success of convolutional neural networks (CNNs) \cite{lecun1998cnn} on the ImageNet \cite{deng2009imagenet}
has attracted considerable attentions in the computer vision research community. Since the publication of the winning model~\cite{krizhevsky2012} of the ImageNet 2012 challenge, CNN-based approaches have been shown to achieve state-of-the-art on many challenging image datasets. 
For instance, 
the trained model in \cite{krizhevsky2012} has
been successfully transferred to the PASCAL VOC dataset and used as a mid-level image representation~\cite{girshick2014rcnn}. 
They showed promising results in object and action classification and localization~\cite{oquab2014}. An evaluation of off-the-shelf CNN features applied to visual classification and visual instance retrieval has been conducted on several image datasets~\cite{razavian2014}. 
The main observation was that the CNN-based approaches outperformed the approaches based on the most successful hand-designed features. 

Recently, CNN architectures trained on videos have emerged, with the objective of capturing and encoding motion information. The 3D CNN model proposed in~\cite{ji2013} outperformed baseline methods in human action recognition. 
The two-stream convolutional network proposed in~\cite{karen-flow} combined a CNN model trained on appearance frames with a CNN model trained on stacked optical flow features to match the performance of hand-crafted spatiotemporal features.
The CNN and long short term memory (LSTM) model have been utilized in~\cite{Ng2015LSTM} to obtain video-level representation.
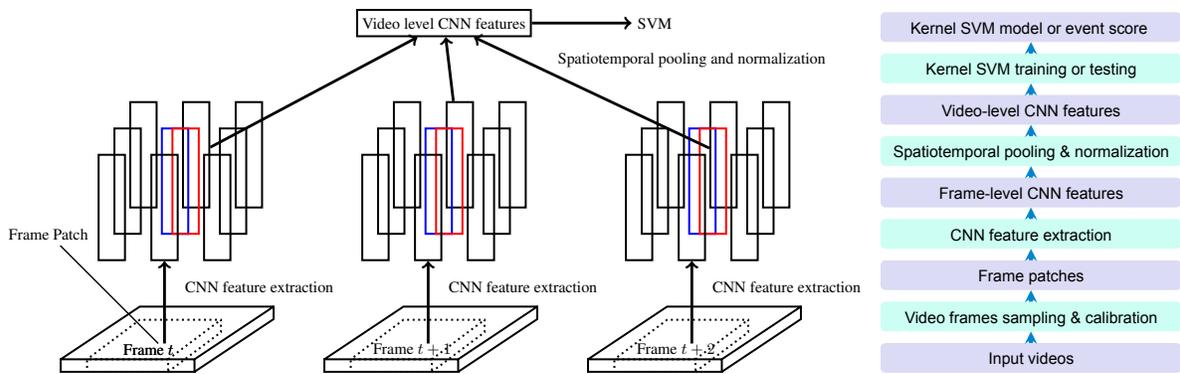
\begin{figure*}[t]
\small
\centering
\hspace{-0.4in}
\resizebox{0.75\linewidth}{!}{%
\makeatletter
\pgfkeys{/pgf/.cd,
  parallelepiped offset x/.initial=2mm,
  parallelepiped offset y/.initial=2mm
}
\pgfdeclareshape{parallelepiped}
{
  \inheritsavedanchors[from=rectangle] 
  \inheritanchorborder[from=rectangle]
  \inheritanchor[from=rectangle]{north}
  \inheritanchor[from=rectangle]{north west}
  \inheritanchor[from=rectangle]{north east}
  \inheritanchor[from=rectangle]{center}
  \inheritanchor[from=rectangle]{west}
  \inheritanchor[from=rectangle]{east}
  \inheritanchor[from=rectangle]{mid}
  \inheritanchor[from=rectangle]{mid west}
  \inheritanchor[from=rectangle]{mid east}
  \inheritanchor[from=rectangle]{base}
  \inheritanchor[from=rectangle]{base west}
  \inheritanchor[from=rectangle]{base east}
  \inheritanchor[from=rectangle]{south}
  \inheritanchor[from=rectangle]{south west}
  \inheritanchor[from=rectangle]{south east}
  \backgroundpath{
    \southwest \pgf@xa=\pgf@x \pgf@ya=\pgf@y
    \northeast \pgf@xb=\pgf@x \pgf@yb=\pgf@y
    \pgfmathsetlength\pgfutil@tempdima{\pgfkeysvalueof{/pgf/parallelepiped
offset x}}
    \pgfmathsetlength\pgfutil@tempdimb{\pgfkeysvalueof{/pgf/parallelepiped
offset y}}
    \def\ppd@offset{\pgfpoint{\pgfutil@tempdima}{\pgfutil@tempdimb}}
    \pgfpathmoveto{\pgfqpoint{\pgf@xa}{\pgf@ya}}
    \pgfpathlineto{\pgfqpoint{\pgf@xb}{\pgf@ya}}
    \pgfpathlineto{\pgfqpoint{\pgf@xb}{\pgf@yb}}
    \pgfpathlineto{\pgfqpoint{\pgf@xa}{\pgf@yb}}
    \pgfpathclose
    \pgfpathmoveto{\pgfqpoint{\pgf@xb}{\pgf@ya}}
    \pgfpathlineto{\pgfpointadd{\pgfpoint{\pgf@xb}{\pgf@ya}}{\ppd@offset}}
    \pgfpathlineto{\pgfpointadd{\pgfpoint{\pgf@xb}{\pgf@yb}}{\ppd@offset}}
    \pgfpathlineto{\pgfpointadd{\pgfpoint{\pgf@xa}{\pgf@yb}}{\ppd@offset}}
    \pgfpathlineto{\pgfqpoint{\pgf@xa}{\pgf@yb}}
    \pgfpathmoveto{\pgfqpoint{\pgf@xb}{\pgf@yb}}
    \pgfpathlineto{\pgfpointadd{\pgfpoint{\pgf@xb}{\pgf@yb}}{\ppd@offset}}
  }
}
\makeatother

\ifx\du\undefined
  \newlength{\du}
\fi
\setlength{\du}{1.5\unitlength}
\ifx\spacing\undefined
  \newlength{\spacing}
\fi
\setlength{\spacing}{30\unitlength}
\begin{tikzpicture}
\pgfsetlinewidth{0.7500\du}
\pgfsetmiterjoin
\pgfsetbuttcap

\node[parallelepiped,draw=black,
  minimum width=50\du,minimum height=\du,
  parallelepiped offset x=30\du,
  parallelepiped offset y=20\du] (1a) at (0, -\spacing) {};
  \node[anchor=center] at (0.4\spacing, -0.7\spacing){Frame $t$};

\node[parallelepiped,draw=black, dotted,
  minimum width=30\du,minimum height=\du,
  parallelepiped offset x=20\du,
  parallelepiped offset y=15\du] (2a) at (0, -\spacing) {};

\node[anchor=center] at (0.4\spacing, -0.7\spacing){Frame $t$};

\node[parallelepiped,draw=black,
  minimum width=50\du,minimum height=\du,
  parallelepiped offset x=30\du,
  parallelepiped offset y=20\du] (1b) at (5\spacing, -\spacing) {};
  \node[anchor=center] at (5.4\spacing, -0.7\spacing){Frame $t+1$};

\node[parallelepiped,draw=black, dotted,
  minimum width=30\du,minimum height=\du,
  parallelepiped offset x=20\du,
  parallelepiped offset y=15\du] (2b) at (5\spacing, -\spacing) {};

\node[parallelepiped,draw=black,
  minimum width=50\du,minimum height=\du,
  parallelepiped offset x=30\du,
  parallelepiped offset y=20\du] (1c) at (10\spacing, -\spacing) {};
  \node[anchor=center] at (10.4\spacing, -0.7\spacing){Frame $t+2$};

\node[parallelepiped,draw=black, dotted,
  minimum width=30\du,minimum height=\du,
  parallelepiped offset x=20\du,
  parallelepiped offset y=15\du] (2c) at (10\spacing, -\spacing) {};

\node[rectangle, draw=black, minimum width=10\du, minimum height=40\du] (f1) at (0.3\spacing, 3.0\spacing) {};
\node[rectangle, draw=black, minimum width=10\du, minimum height=40\du] (f2) at (1.3\spacing, 3.0\spacing) {};
\node[rectangle, draw=black, minimum width=10\du, minimum height=40\du] (f3) at (2.3\spacing, 3.0\spacing) {};

\node[rectangle, draw=black, minimum width=10\du, minimum height=40\du] (f4) at (0\spacing, 2.5\spacing) {};
\node[rectangle, draw=black, minimum width=10\du, minimum height=40\du, color=blue] (f5) at (0.9\spacing, 2.5\spacing) {};
\node[rectangle, draw=black, minimum width=10\du, minimum height=40\du, color=red] (f10) at (1.1\spacing, 2.5\spacing) {};
\node[rectangle, draw=black, minimum width=10\du, minimum height=40\du] (f6) at (2\spacing, 2.5\spacing) {};

\node[rectangle, draw=black, minimum width=10\du, minimum height=40\du] (f7) at (-0.3\spacing, 2.0\spacing) {};
\node[rectangle, draw=black, minimum width=10\du, minimum height=40\du] (f8) at (0.7\spacing, 2.0\spacing) {};
\node[rectangle, draw=black, minimum width=10\du, minimum height=40\du] (f9) at (1.7\spacing, 2.0\spacing) {};

\node[rectangle, draw=black, minimum width=10\du, minimum height=40\du] (g1) at (5.3\spacing, 3.0\spacing) {};
\node[rectangle, draw=black, minimum width=10\du, minimum height=40\du] (g2) at (6.3\spacing, 3.0\spacing) {};
\node[rectangle, draw=black, minimum width=10\du, minimum height=40\du] (g3) at (7.3\spacing, 3.0\spacing) {};

\node[rectangle, draw=black, minimum width=10\du, minimum height=40\du] (g4) at (5\spacing, 2.5\spacing) {};
\node[rectangle, draw=black, minimum width=10\du, minimum height=40\du, color=blue] (g5) at (5.9\spacing, 2.5\spacing) {};
\node[rectangle, draw=black, minimum width=10\du, minimum height=40\du, color=red] (g10) at (6.1\spacing, 2.5\spacing) {};
\node[rectangle, draw=black, minimum width=10\du, minimum height=40\du] (g6) at (7\spacing, 2.5\spacing) {};

\node[rectangle, draw=black, minimum width=10\du, minimum height=40\du] (g7) at (4.7\spacing, 2.0\spacing) {};
\node[rectangle, draw=black, minimum width=10\du, minimum height=40\du] (g8) at (5.7\spacing, 2.0\spacing) {};
\node[rectangle, draw=black, minimum width=10\du, minimum height=40\du] (g9) at (6.7\spacing, 2.0\spacing) {};

\node[rectangle, draw=black, minimum width=10\du, minimum height=40\du] (h1) at (10.3\spacing, 3.0\spacing) {};
\node[rectangle, draw=black, minimum width=10\du, minimum height=40\du] (h2) at (11.3\spacing, 3.0\spacing) {};
\node[rectangle, draw=black, minimum width=10\du, minimum height=40\du] (h3) at (12.3\spacing, 3.0\spacing) {};

\node[rectangle, draw=black, minimum width=10\du, minimum height=40\du] (h4) at (10\spacing, 2.5\spacing) {};
\node[rectangle, draw=black, minimum width=10\du, minimum height=40\du, color=blue] (g5) at (10.9\spacing, 2.5\spacing) {};
\node[rectangle, draw=black, minimum width=10\du, minimum height=40\du, color=red] (g10) at (11.1\spacing, 2.5\spacing) {};
\node[rectangle, draw=black, minimum width=10\du, minimum height=40\du] (h6) at (12\spacing, 2.5\spacing) {};

\node[rectangle, draw=black, minimum width=10\du, minimum height=40\du] (h7) at (9.7\spacing, 2.0\spacing) {};
\node[rectangle, draw=black, minimum width=10\du, minimum height=40\du] (h8) at (10.7\spacing, 2.0\spacing) {};
\node[rectangle, draw=black, minimum width=10\du, minimum height=40\du] (h9) at (11.7\spacing, 2.0\spacing) {};

\node[anchor=center] (1ap) at (0.7\spacing, -0.7\spacing){};
\node[anchor=center] (2ap) at (5.7\spacing, -0.7\spacing){};
\node[anchor=center] (3ap) at (10.7\spacing, -0.7\spacing){};
\node[anchor=center] (1aq) at (-1.5\spacing, 1.5\spacing){Frame Patch};
\draw[ultra thick, ->] (1ap) -- (f8);
\draw[ultra thick, ->] (2ap) -- (g8);
\draw[ultra thick, ->] (3ap) -- (h8);
\draw[thick, -] (1aq) -- (1ap);
\node[rectangle, draw=black, minimum width=60\du, minimum height=10\du] (pool) at (6\spacing, 5.5\spacing) {Video level CNN features};

\node[anchor=center] at (2.5\spacing, 0.5\spacing){CNN feature extraction};
\node[anchor=center] at (7.5\spacing, 0.5\spacing){CNN feature extraction};
\node[anchor=center] at (12.5\spacing, 0.5\spacing){CNN feature extraction};
\node[anchor=center] at (10.7\spacing, 4.8\spacing){Spatiotemporal pooling and normalization};
\node[anchor=center] (svm) at (10\spacing, 5.5\spacing){SVM};
\draw[ultra thick, ->] (f2) -- (pool);
\draw[ultra thick, ->] (g2) -- (pool);
\draw[ultra thick, ->] (h2) -- (pool);
\draw[ultra thick, ->] (pool) -- (svm);
\end{tikzpicture}
}
\includegraphics[width=0.23\linewidth]{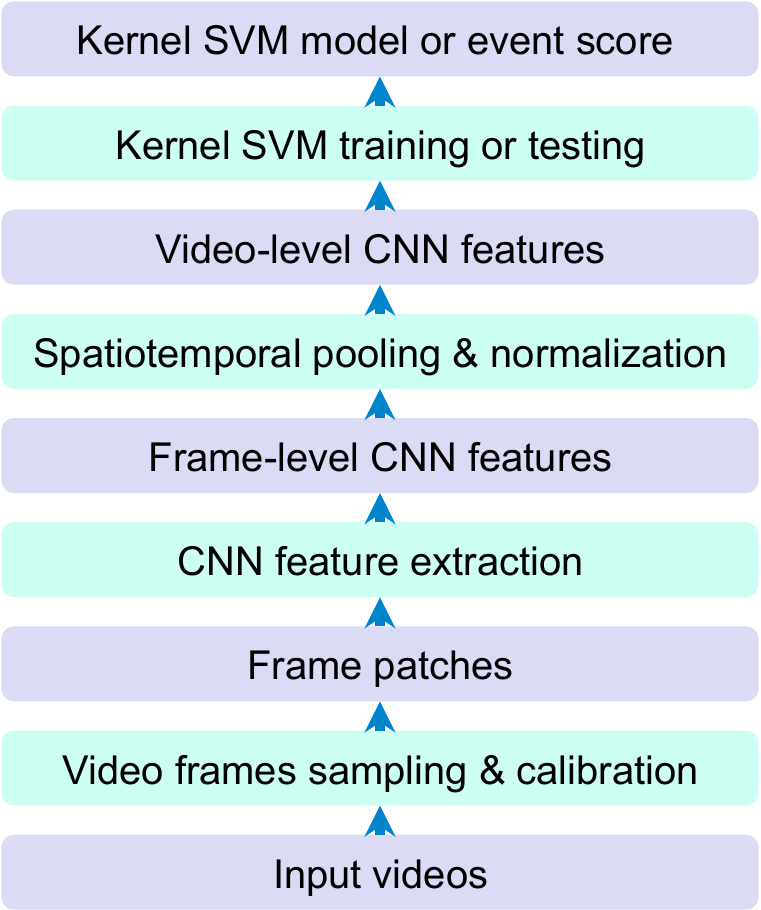}
\caption{Overview of the proposed video classification pipeline.}
\label{fig:overview}
\end{figure*}

In this paper, we propose an efficient approach to exploit off-the-shelf \emph{image-trained} CNN architectures for video classification (Figure~\ref{fig:overview}).
Our contributions are the following:
\begin{itemize}[noitemsep,nolistsep] 
  \item We discuss each step of the proposed video classification pipeline, including the choice of CNN layers, the video frame sampling and calibration, the spatial and temporal pooling, the feature normalization, and the choice of classifier. All our design choices are supported by extensive experiments on the TRECVID MED'14 video dataset.
  \item We provide thorough experimental comparisons between the CNN-based approach and some state-of-the-art static and motion-based approaches, showing that the CNN-based approach can outperform the latter, both in terms of accuracy and speed. 
  \item We show that integrating motion information with a simple average fusion considerably improves classification performance, achieving the state-of-the-art performance on TRECVID MED'14 and UCF-101.
\end{itemize} 
Our work is closely related to other research efforts towards the efficient use of CNN for video classification. While it is now clear that CNN-based approaches outperform most state-of-the-art handcrafted features for image classification~\cite{razavian2014}, it is not yet obvious that this holds true for video classification. Moreover, there seems to be mixed conclusions regarding the benefit of training a spatiotemporal vs. applying an image-trained CNN architecture on videos. 
Indeed, while Ji~\emph{et al.}~\cite{ji2013} observed a significant gain using 3D convolutions over the 2D CNN architectures 
and Simonyan~\emph{et al.}~\cite{karen-flow} obtained substantial gains over an appearance based 2D CNN using optical flow features alone, 
Karpathy~\emph{et al.}~\cite{karpathy2014} reported only a moderate improvement. 
Although the specificity of the considered video datasets might play a role, the way the 2D CNN architecture is exploited for video classification is certainly the main reason behind these contradictory observations. The additional computational cost of training on videos is also an element that should be taken into account when comparing the two options. Prior to training a spatiotemporal CNN architecture, it thus seems legitimate to fully exploit the potential of image-trained CNN architectures. Obtained on a highly heterogeneous video dataset, we believe that our results can serve as a strong 2D CNN baseline against which to compare CNN architectures specifically trained on videos.      

\section{Deep Convolutional Neural Networks}
\label{sec:dcnn}
Convolutional neural networks~\cite{lecun1998cnn} consist of layers of spatially-structured hidden units. 
Each hidden unit typically looks at a small patch of hidden (or input)
units in the previous layer, 
applies convolution or pooling operations to it and then 
non-linearity to the result to compute its own state. 
A spatial patch of units is convolved with multiple filters (learned weights) to generate
feature maps. 
A pooling operation takes a spatial patch and computes 
typically maximum or average activation (max or average pooling)
of that patch
for each input channel,
creating translation invariance.
At the top of the stacked layers, 
the spatially organized hidden units are usually densely connected, 
which eventually connect to the output units (e.g. softmax for classification). 
Regularizers, such as $\ell_2$ decay and dropout~\cite{dropout}, have been shown to be effective for preventing overfitting.
The structure of deep neural nets enables the hidden layers to learn rich
distributed representations of the input images.
The densely connected layers,
in particular, 
can be seen as learning a high-level representation of the image
accumulating information from all the spatial locations.

We initially developed our work on the CNN architecture by Krizhevsky \textit{et al.}~\cite{krizhevsky2012}.
In the recent ILSVRC-2014 competition,
the top performing models GoogLeNet~\cite{googlenet} and VGG~\cite{oxford_vdeep} 
used smaller receptive fields 
and increased depth, 
showing superior performance over~\cite{krizhevsky2012}.
We adopt the publicly available pretrained VGG model 
for its superior post-competition single model performance over GoogLeNet
and the popular Krizhevsky's model in our system.
The proposed video classification approach is generic with respect to the CNN architectures,
therefore, can be adapted to other CNN architectures as well.
\section{Video Classification Pipeline}
\label{sec:pipeline}
Figure~\ref{fig:overview} gives an overview of the proposed video classification pipeline.
Each component of the pipeline is discussed in this sections.

\begin{figure*}[t]
\newcommand{\mysize}{0.33}
\small
\begin{center}
\includegraphics[width=\mysize\linewidth]{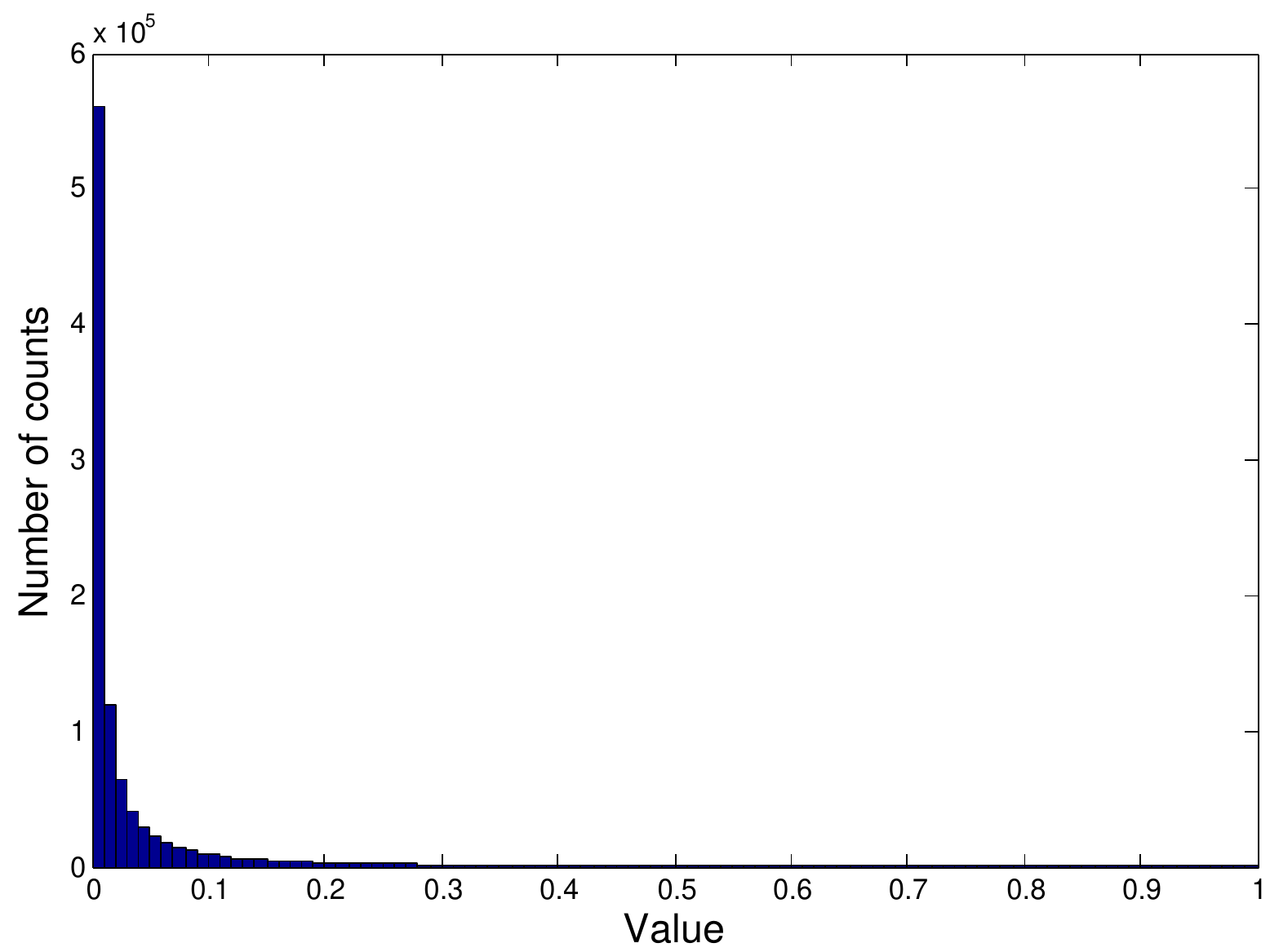}
\hfill
\includegraphics[width=\mysize\linewidth]{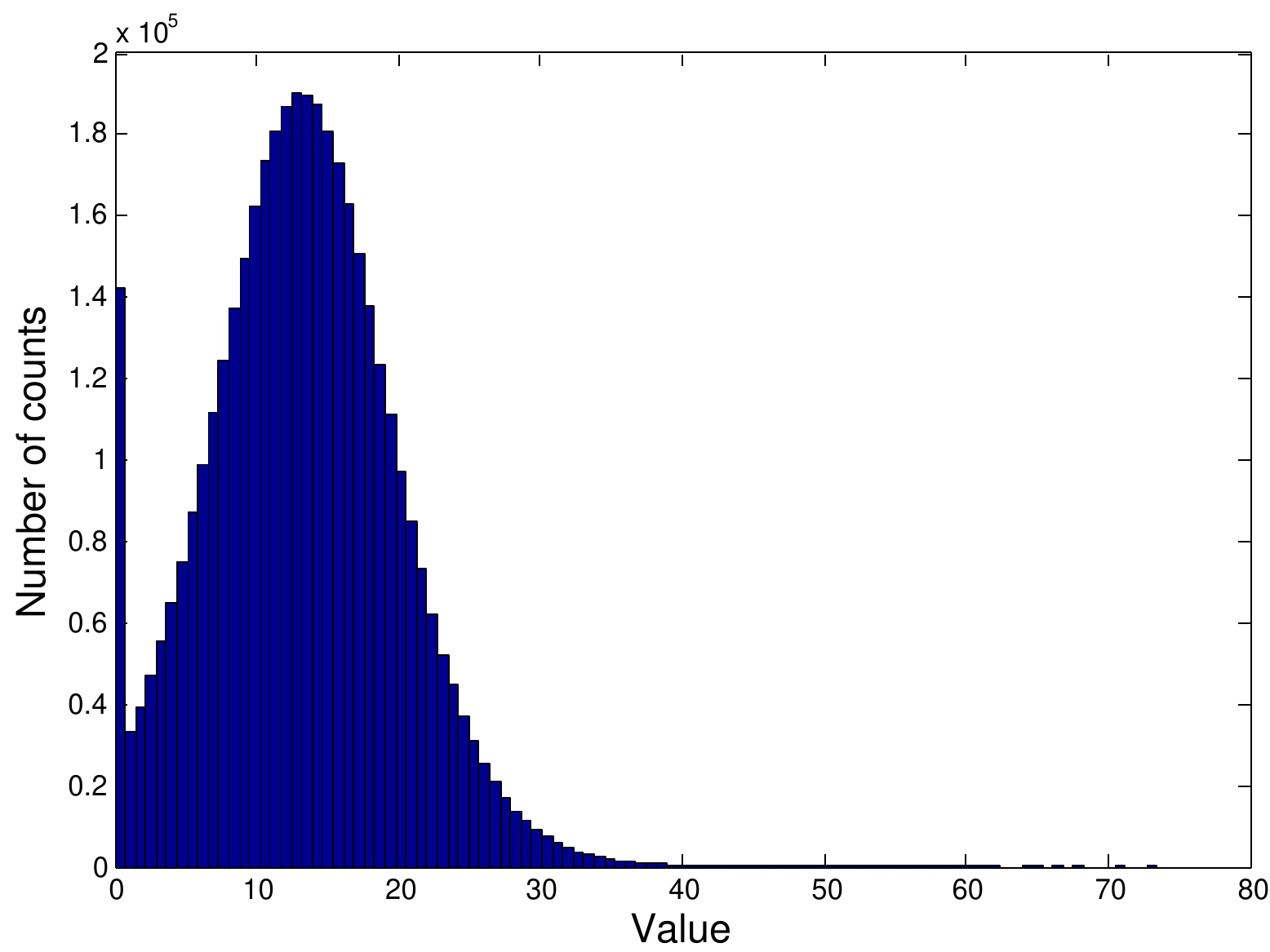}
\hfill
\includegraphics[width=\mysize\linewidth]{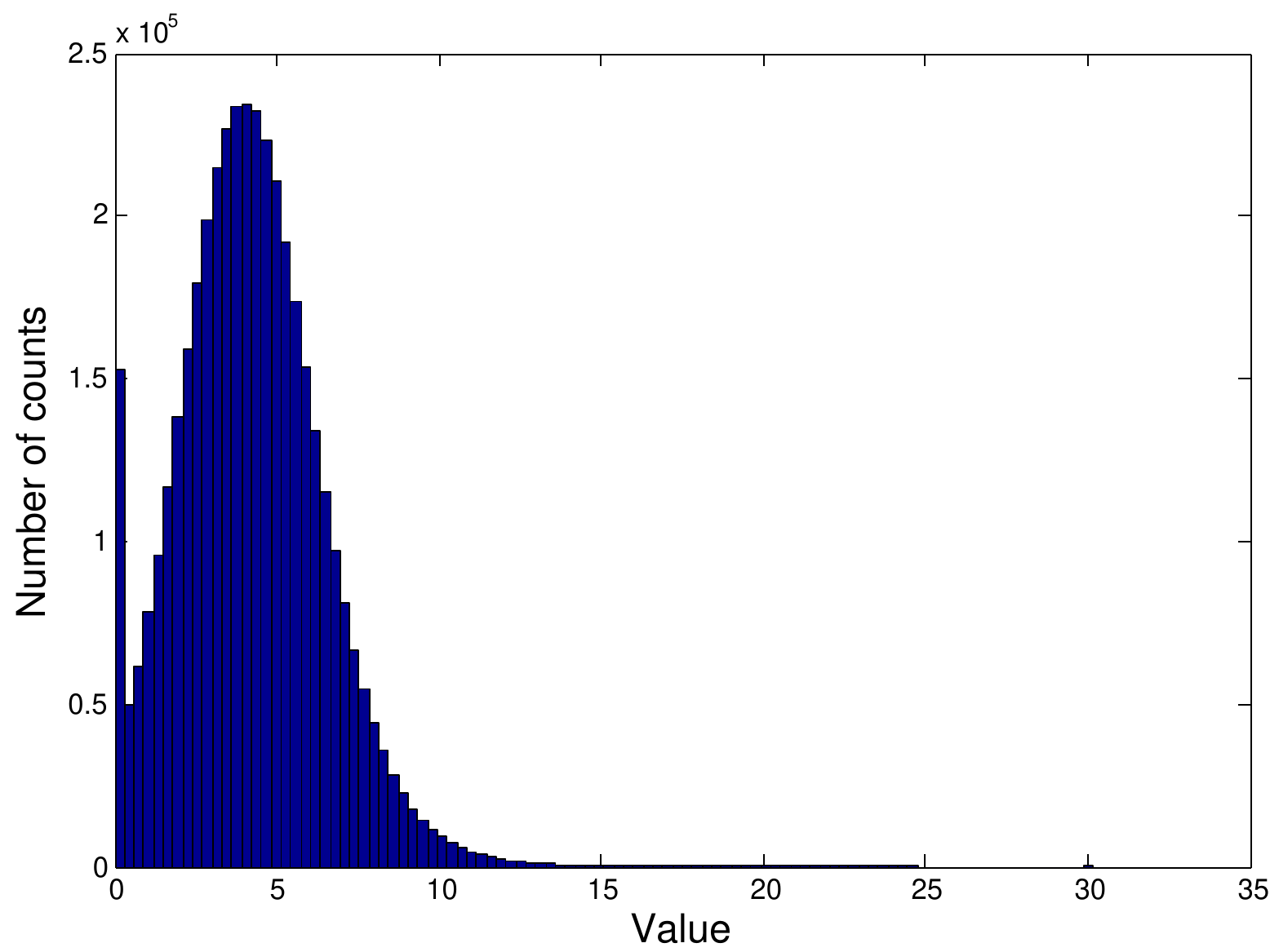}
\end{center}
\caption{Distribution of values for the output layer and the last two hidden layers of the CNN architecture. 
Left:~output. Middle:~hidden6. Right:~hidden7.}\label{fig:out_fc6_fc7}
\end{figure*}

\textbf{Choice of CNN Layer}
We have considered the output layer and the last two hidden layers (fully-connected layers) as CNN-based features. The 1,000-dimensional output-layer features, with values between $[0,1]$, are the posterior probability scores corresponding to the 1,000 classes from the ImageNet dataset. Since our events of interest are different from the 1,000 classes, the output-layer features are rather sparse (see Figure~\ref{fig:out_fc6_fc7} left panel).
The hidden-layer features are treated as high-level image representations in our video classification pipeline. These features are outputs of rectified linear units (RELUs). Therefore, they are lower bounded by zero but do not have a pre-defined upper bound (see Figure~\ref{fig:out_fc6_fc7}~middle and right panels) and thus require some normalization. 


\textbf{Video Frame Sampling and Calibration}
We uniformly sampled 50 to 120 frames depending on the clip length.
We have explored alternative frame sampling schemes (e.g., based on keyframe detection), 
but found that they all essentially yield the same performance as uniform sampling. 
The first layer of the CNN architecture takes $224\times224$ RGB images as inputs. 
We extracted multiple patches from the frames and rescale each of them to a size of $224\times224$. 
The number, location and original size of the patches are determined by selected spatial pooling strategies.
Each of $P$ patches is then used as input to the CNN architecture,
providing $P$ different outputs for each video frame.

\begin{figure*}[t]
\small
\newcommand{\mysizeA}{0.138}
\newcommand{\mysizeB}{0.408}
\newcommand{\mysizeC}{0.44}
\begin{center}
\begin{minipage}[b]{\mysizeA\linewidth}
\includegraphics[width=\linewidth]{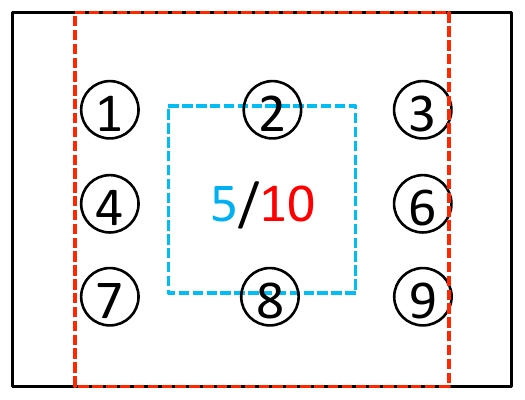}
\end{minipage}
\hfill
\begin{minipage}[b]{\mysizeB\linewidth}
\includegraphics[width=\linewidth]{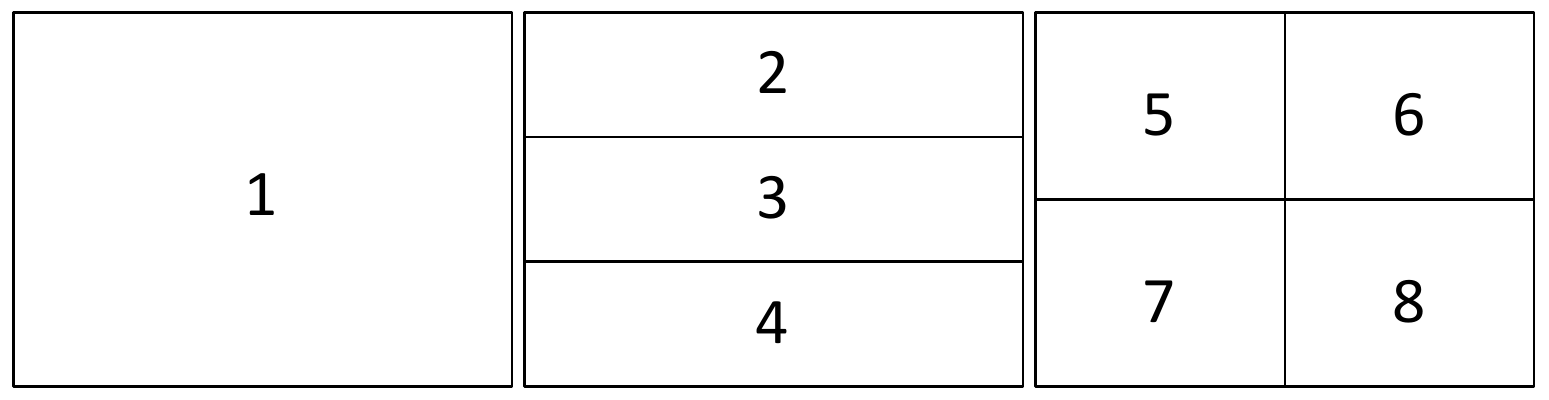}
\end{minipage}
\hfill
\begin{minipage}[b]{\mysizeC\linewidth}
\newcommand{\mysize}{0.323}
  \begin{center}
    \includegraphics[width=\mysize\linewidth]{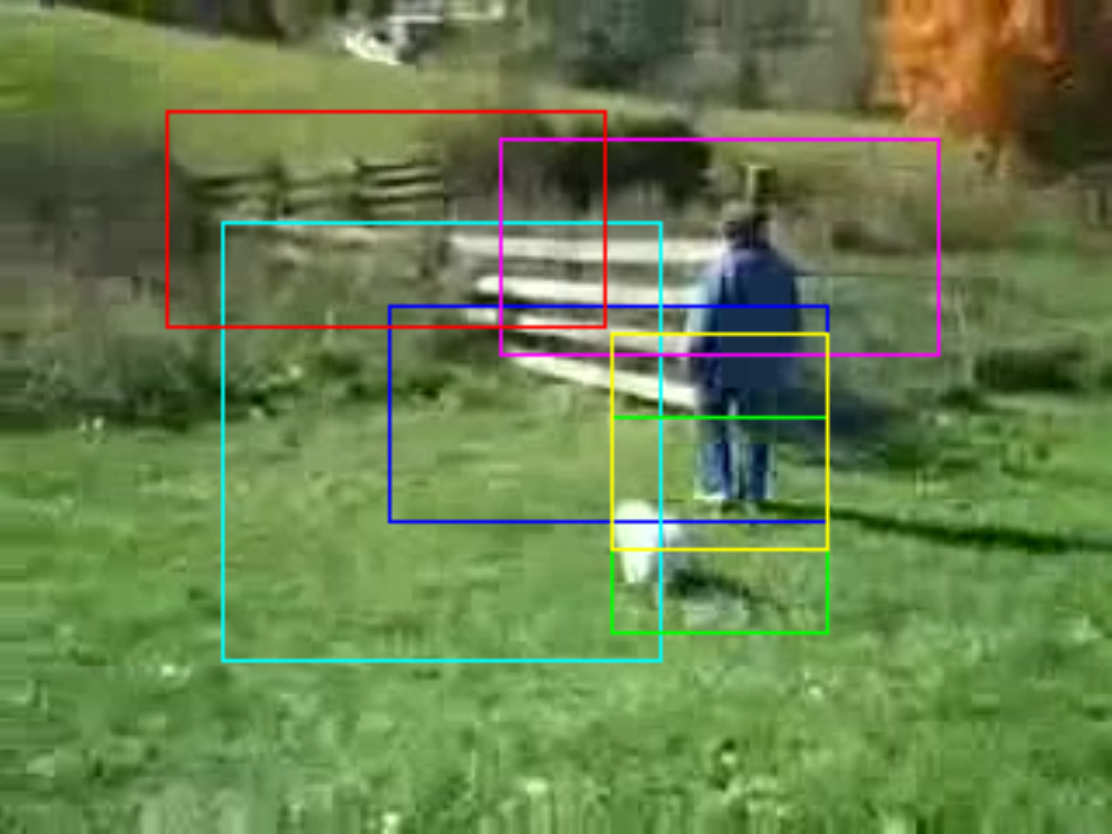}  \hfill
    \includegraphics[width=\mysize\linewidth]{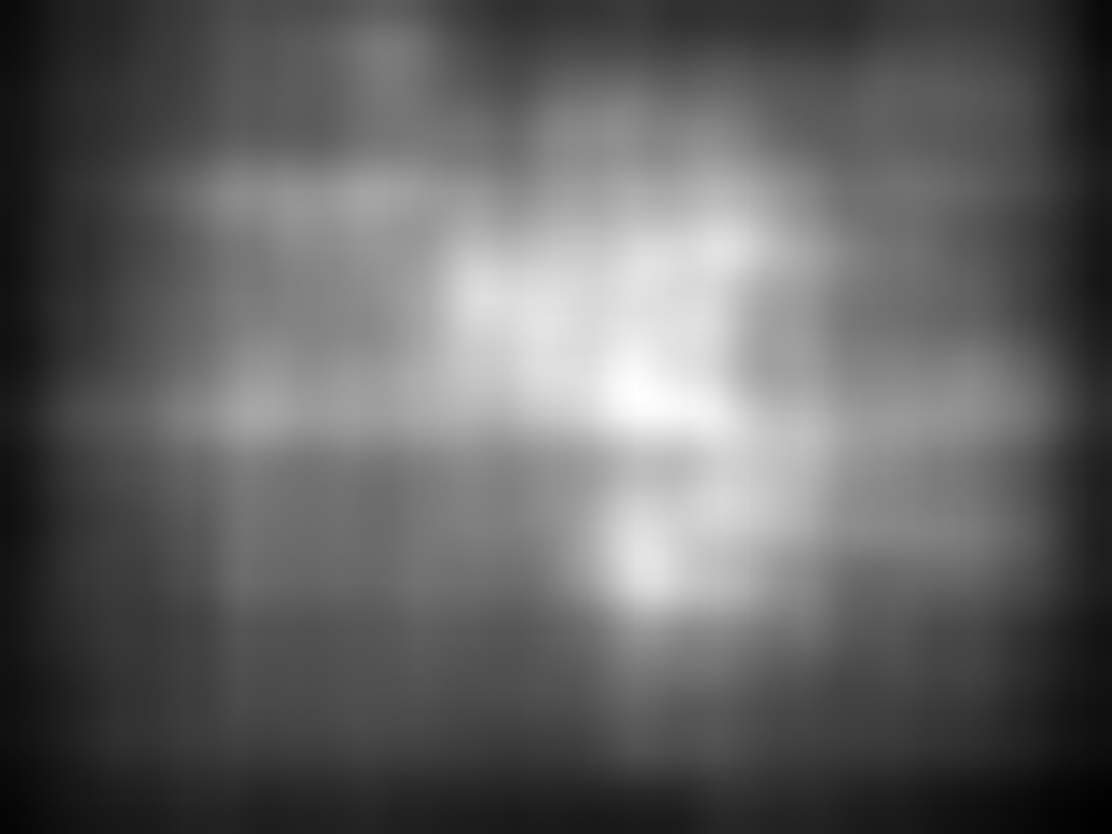}  \hfill
    \includegraphics[width=\mysize\linewidth]{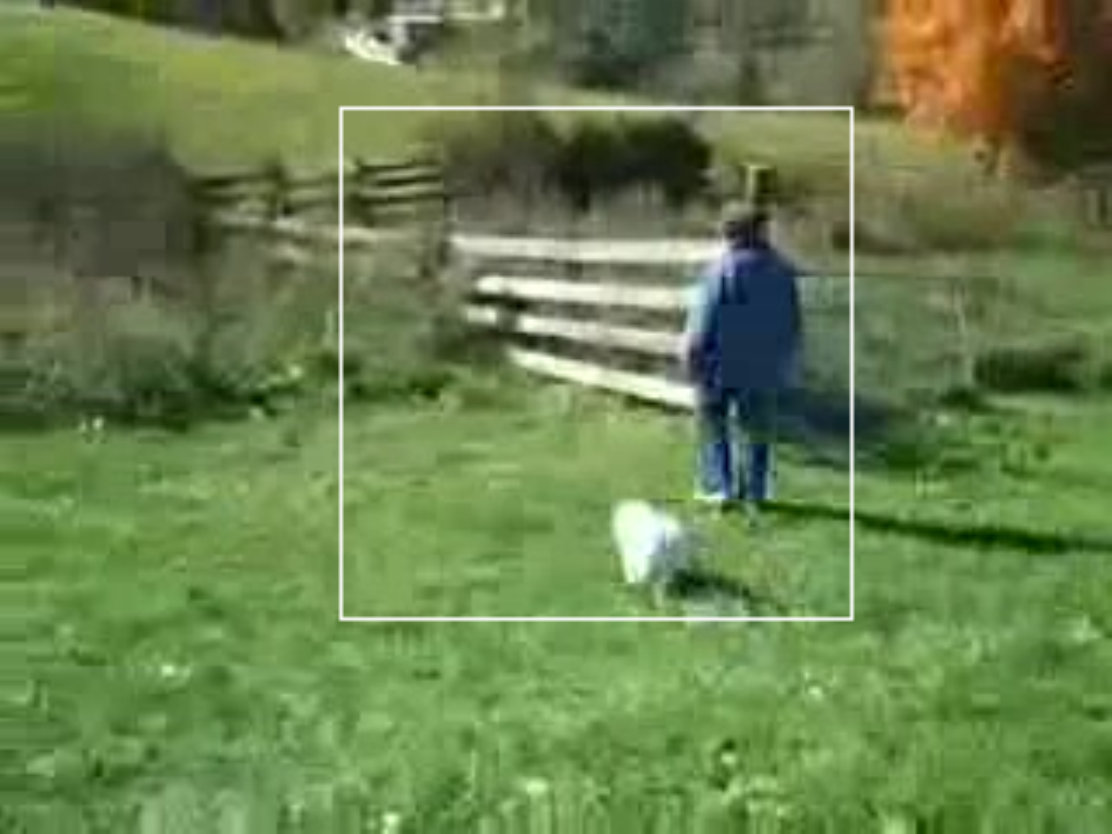}
  \end{center}
\end{minipage}
\\
\begin{minipage}[b]{\mysizeA\linewidth}\centering
(a)
\end{minipage}
\hfill
\begin{minipage}[b]{\mysizeB\linewidth}\centering
(b)
\end{minipage}
\begin{minipage}[b]{\mysizeC\linewidth}\centering
(c)
\end{minipage}
\end{center}
\caption{Pooling. (a). The 10 overlapping square patches extracted from each frame for spatial pyramid pooling. The red patch (centered at location 10) has sides equal to the frame height, whereas the other patches (centered at locations 1-9) have sides equal to half of the frame height.
(b).
Spatial pyramids (SP). Left: the $1 \times 1$ spatial partition includes the entire frame. Middle: the $3 \times 1$ spatial partitions include the top, middle and bottom of the frame. Right: the $2 \times 2$ spatial partitions include the upper-left, upper-right, bottom-left and bottom-right of the frame.
(c). 
Objectness-based pooling. Left: objectness proposals. Middle: sum of 1000 objectness proposals. Right: foreground patch from thresholding proposal sums.
}
\label{fig:pooling}
\end{figure*}

\textbf{Spatial and Temporal Pooling}
We evaluated two pooling schemes, average and max, for both spatial and temporal pooling along with two spatial partition schemes.
Inspired by the \textit{spatial pyramid} approach developed for bags of features~\cite{lazebnik2006}, we spatially pooled together the CNN features computed on patches centered in the same pre-defined sub-region of the video frames. 
As illustrated in Figure~\ref{fig:pooling}, 10 overlapping square patches are extracted from each frame.
We have considered 8 different regions consisting of $1\times1$, $3\times1$ and $2\times2$ partitions of the video frame. 
For instance, features from patches 1, 2, 3 will be pooled together to yield a single feature in region 2.
Such region specification considers the full frame, the vertical structure and four corners of a frame.
The pooled features from 8 spatial regions are concatenated into one feature vector.
Our spatial pyramid approach differs from that of~\cite{eccv2014he,gong2014multi,DBLP:journals/corr/XuYH14} in that 
it is applied to the output- or hidden-layer feature response of multiple inputs
rather than between the convolutional layer and fully-connected layer. 
The pre-specified regions are also different.
We have also considered utilizing \textit{objectness} to guide feature pooling
by concatenating the CNN features extracted from the foreground region and the full frame.
As shown in Figure~\ref{fig:pooling},
the foreground region is resulted from thresholding the sum of 1000 objectness proposals generated from BING~\cite{cheng2014bing}.
Unlike R-CNN~\cite{girshick2014rcnn} that extract CNN features from all objectness proposals for object detection and localization, 
We used only one coarse foreground region to reduce the computational cost.


\textbf{Feature Normalization}
We compared different normalization schemes and identified normalization most appropriate for the particular feature type.
Let ${\bf f}\in\mathbb{R}^D$ denote the original video-level CNN feature vector and ${\bf\tilde{f}}\in\mathbb{R}^D$ its normalized version. We have investigated three different normalizations:
$\ell_1$ normalization ${\bf\tilde{f}}={\bf f}/\Vert{\bf f}\Vert_1$; 
$\ell_2$ normalization ${\bf\tilde{f}}={\bf f}/\Vert{\bf f}\Vert_2$, 
which is typically performed prior to training an SVM model; 
and
\emph{root} normalization ${\bf\tilde{f}}=\sqrt{{\bf f}/\Vert{\bf f}\Vert_1}$
introduced in~\cite{arandjelovic2012} and shown to improve the performance of SIFT descriptors.  

\textbf{Choice of Classifier}
We applied SVM classifiers to the video-level features, including linear SVM and non-linear SVM 
with Gaussian radial basis function (RBF) kernel $\exp \{-\gamma || \mathbf{x} - \mathbf{y} ||^2 \}$
and
exponential $\chi^2$ kernel $\exp\{-\gamma \sum_i{\frac{(x_i - y_i)^2}{x_i+y_i}}\}$.
Principal component analysis (PCA) can be applied prior to SVM
to reduce feature dimensions.
\section{Modality Fusion}
\subsection{Fisher Vectors}
\label{ssec:fv}
An effective approach to video classification is to
extract multiple low-level feature descriptors; 
and then encode them as a fixed-length video-level Fisher vector (FV). 
The FV is a generalization of the bag-of-words approach that encodes the zero-order, the first- and the second-order statistics of the descriptors distribution.
The FV encoding procedure is summarized as follows~\cite{perronnin2010}.
First, learn a Gaussian mixture model (GMM) on low-level descriptors extracted from a generic set of unlabeled videos.
Second, compute the gradients of the log-likelihood of the GMM (known as the score function) with respect to the GMM parameters. The gradient of the score function with respect to the mixture weight parameters encodes the zero-order statistics. The gradient with respect to the Gaussian means encodes the first-order statistics, while the gradient with respect to the Gaussian variance encodes the second-order statistics. 
Third, concatenate the score function gradients into a single vector and apply a signed square rooting on each FV dimension (power normalization) followed by a global $\ell_2$ normalization.
As low-level features, we considered both the standard D-SIFT descriptors~\cite{lowe2004} and the more sophisticated motion-based improved dense trajectories (IDT)~\cite{wang2013action}. 
For the SIFT descriptors, we opted for multiscale (5 scales) and dense (stride of 4 pixels in both spatial dimensions) sampling, root normalization and spatiotemporal pyramid pooling. 
For the IDT descriptors, we concatenated HOG~\cite{dalal2005}, HOF~\cite{dalal2006} and MBH~\cite{laptev2008} descriptors extracted along the estimated motion trajectory.
\subsection{Fusion}
\label{ssec:fusion}
We investigated modality fusion to fully utilize the CNN features and strong the hand-engineered features.
The weighted average fusion was applied:
first, the SVM margins was converted into posterior probability scores using Platt's method~\cite{platt1999};
and then the weights of the linear combination of the considered features scores for each class was optimized using cross-validation.
We evaluated fusion features scores of different CNN layers and FVs.

\section{Evaluation in Event Detection}
\subsection{Video Dataset and Performance Metric}
\label{sec:dataset}
\begin{figure*}[t]
\begin{center}
\newcommand{\mydir}{./fig/examples}
\newcommand{\mysize}{0.195}
\newcommand{\mysizeA}{0.195}
\newcommand{\mysizeB}{0.195}
\newcommand{\mysizeC}{0.195}
\small
\begin{minipage}[b]{1.0\linewidth}\centering
    \includegraphics[width=\mysize\linewidth]{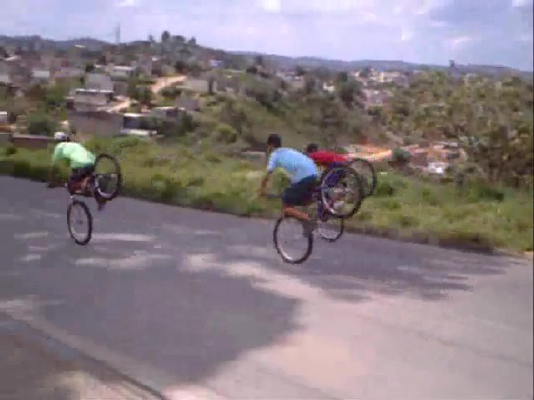}
    \hfill
    \includegraphics[width=\mysize\linewidth]{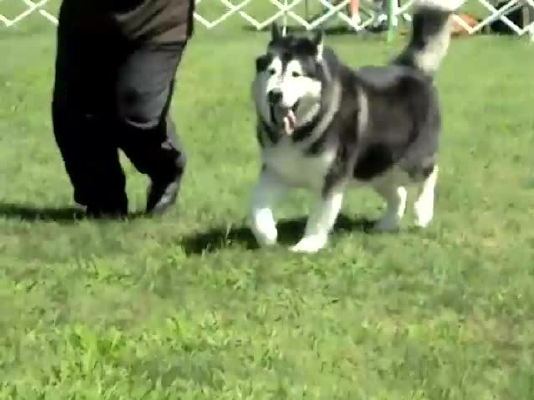}
    \hfill
    \includegraphics[width=\mysize\linewidth]{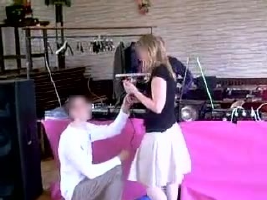}
    \hfill
    \includegraphics[width=\mysize\linewidth]{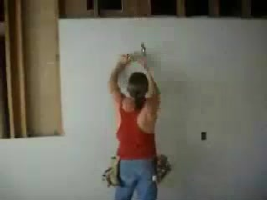}
    \hfill
    \includegraphics[width=\mysize\linewidth]{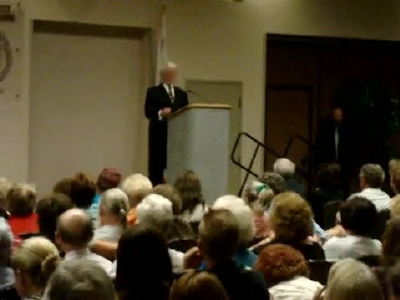}
\end{minipage}
\\
\begin{minipage}[b]{\mysizeA\linewidth}\centering
Attempting Bike trick
\end{minipage}
\begin{minipage}[b]{\mysizeC\linewidth}\centering
Dog show
\end{minipage}
\hfill
\begin{minipage}[b]{\mysizeB\linewidth}\centering
Marriage proposal
\end{minipage}
\hfill
\begin{minipage}[b]{\mysizeB\linewidth}\centering
Renovating a home
\end{minipage}
\hfill
\begin{minipage}[b]{\mysizeB\linewidth}\centering
Town hall meeting
\end{minipage}
\\
\begin{minipage}[b]{1.0\linewidth}\centering
    \includegraphics[width=\mysize\linewidth]{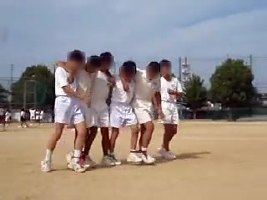}
    \hfill
    \includegraphics[width=\mysize\linewidth]{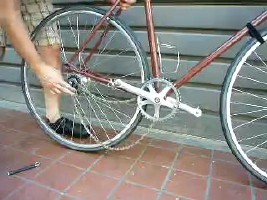}
    \hfill
    \includegraphics[width=\mysize\linewidth]{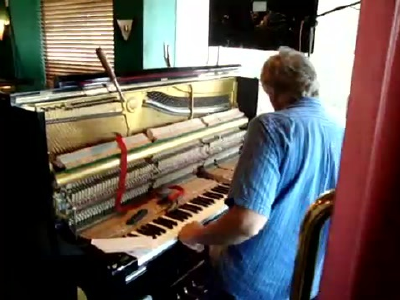}
    \hfill
    \includegraphics[width=\mysize\linewidth]{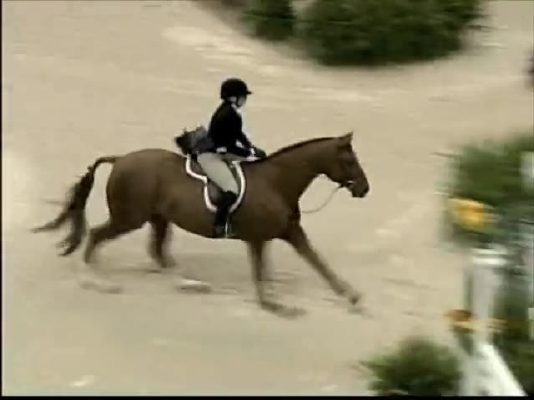}
    \hfill
    \includegraphics[width=\mysize\linewidth]{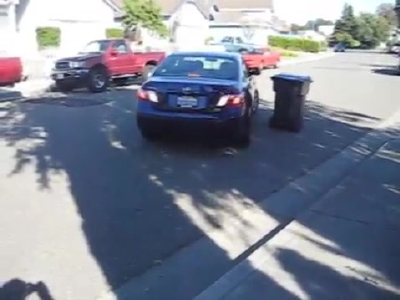}
\end{minipage}
\\
\begin{minipage}[b]{\mysizeB\linewidth}
\center
Winning a race without a vehicle
\end{minipage}
\hfill
\begin{minipage}[b]{\mysizeB\linewidth}
\center
Non-motorized vehicle repair
\end{minipage}
\hfill
\begin{minipage}[b]{\mysizeB\linewidth}
\center
Fixing a musical instrument
\end{minipage}
\hfill
\begin{minipage}[b]{\mysizeB\linewidth}
\center
Horse riding competition
\end{minipage}
\hfill
\begin{minipage}[b]{\mysizeB\linewidth}
\center
Parking a vehicle
\end{minipage}
\caption{Sample frames from the TRECVID MED '14 dataset.}
\label{fig:examples}
\end{center}
\end{figure*}

We conducted extensive experiments on the TRECVID multimedia event detection (MED)'14 video dataset~\cite{over2014}.
This dataset consists of:
(a) a training set of 4,992 unlabeled \emph{background} videos used as the negative examples;
(b) a training set of 2,991 \emph{positive} and \emph{near-miss} videos including 100 positive videos and about 50 near-miss videos (treated as negative examples) for each of the 20 pre-specified events (see Table~\ref{tab:events});
and (c) a test set of 23,953 videos contains positive and negative instances of the pre-specified events.
Some sample frames are given in Figure~\ref{fig:examples}.
Contrary to other popular video datasets, such as UCF-101~\cite{soomro2012}, the MED'14 dataset is not constrained to any class of videos. It consists of a heterogeneous set of temporally untrimmed YouTube-like videos of various resolutions, quality, camera motions, and illumination conditions. This dataset is thus one of the largest and the most challenging dataset for video event detection. 
As a retrieval performance metric, we considered the one used in the official MED'14 task, i.e., mean average precision (mAP) across all events. 
Let $E$ denote the number of events, $P_e$ the number of positive instances of event $e$, then mAP is computed as
\begin{equation}
\textrm{mAP} = \frac{1}{E}\sum^E_{e=1}\textrm{AP}(e)\textrm{,}
\end{equation}
where the average precision of an event $e$ is defined as
\begin{equation}
\textrm{AP}(e) = \frac{1}{P_e}\sum^{P_e}_{tp=1}\frac{tp}{\textrm{rank}(tp)}\textrm{.}\nonumber
\end{equation}
mAP is thus normalized between 0 (low classification performance) and 1 (high classification performance). In this paper, we will report it in percentage value. 
The mAP is normalized between 0 (low classification performance) and 1 (high classification performance). In this paper, we will report it in percentage value. 
\begin{table}
\setlength{\tabcolsep}{3pt}
\small
\begin{center}
\begin{tabular}{|l|l|}
\hline
\textbf{Events E021-E030} & \textbf{Events E031-E040} \\
\hline
Attempting a bike trick & Beekeeping \\
Cleaning an appliance & Wedding shower \\
Dog show & Non-motorized vehicle repair \\
Giving directions & Fixing a musical instrument \\
Marriage proposal & Horse riding competition \\
Renovating a home & Felling a tree \\
Rock climbing & Parking a vehicle \\
Town hall meeting & Playing fetch \\
Winning a race w/o a vehicle & Tailgating \\
Working on a metal crafts project & Tuning a musical instrument \\
\hline
\end{tabular}
\end{center}
\caption{TRECVID 2014 pre-specified events.}
\label{tab:events}
\end{table}

\subsection{Single Feature Performance}
The single feature performance with various configurations are reported in Table
\ref{tab:med14_single_feat}
and \ref{tab:med14_obj}.
The following are a few salient observations.

  \textbf{CNN architectures and layers}
  The deeper CNN architecture yields consistently better performance 
  resulted from the depths and small receptive field in all convolutional layers.
  We also observed that both hidden layers outperformed the output layer
  when the CNN architecture, normalization and spatiotemporal pooling strategies are the same.

  \textbf{Pooling, spatial pyramids and objectness}
  We observed a consistent gain for max pooling over average pooling 
  for both spatial and temporal pooling, irrespectively of the used CNN layer. 
  It is mainly resulted from the highly heterogeneous structure of the video dataset. 
  A lot of videos contain frames that can be considered irrelevant or at least less relevant than others; 
  e.g., introductory text and black frames. 
  Hence it is beneficial to use the maximum features response, instead of giving an equal weight to all features.
%
  As observed in Table~\ref{tab:med14_single_feat}, 
  concatenating the 8 spatial partitions (SP8) gives the best performance 
  in all CNN layers and SVM choices (up to $6\%$ mAP gain over no spatial pooling, i.e., ``none''), 
  but at the expense of an increased feature dimensionality and consequently, 
  an increased training and testing time. 
%
  Alternatively, objectness-guided pooling provides a good trade-off between performance and dimensionality,
  shown in Table~\ref{tab:med14_obj}. 
  It outperforms the baseline approach without spatial expansion (``none'' in Table~\ref{tab:med14_single_feat}).
  The feature dimensions are only one-fourth of that of SP8;
  while the performance nearly matches that of SP8 using kernel SVM.

  \textbf{Normalization}
  We observed that $\ell_1$ normalization did not perform well;
  while the $\ell_2$ normalization is essential to achieve good performance
  with hidden layers.
  Applying root normalization to the output layer 
  yields essentially the same result as applying $\ell_2$.
  Yet, we noticed a drop in performance when the root normalization was applied to the hidden layers. 

  \textbf{Classifier}
  One SVM model was trained for each event following the TRECVID MED'14 training rules; 
  i.e., excluding the positive and near-miss examples of the other events. 
  We observed that kernel SVM consistently outperformed linear SVM regardless of CNN layer 
  or normalization. 
  For the output layer, which essentially behaves like a histogram-based feature, 
  the $\chi^2$ kernel yields essentially the same result as RBF kernel.
  For the hidden layers, the best performance was obtained with a RBF kernel. 
  As shown in Table~\ref{tab:med14_single_feat},
  the result of kernel SVM largely outperforms that of linear SVM.

  \textbf{PCA}
  We analyzed dimensionality reduction of the top performing feature, CNN-B-hidden7-SP8,
  shown in Table~\ref{tab:pca}. 
  Applying PCA reduced the feature dimension from $32,768$ to $4096$, $2048$ and $1024$ prior to SVM,
  which resulted in slightly reduced mAPs as compared to classification of features without PCA.
  However, it still outperformed or matched the performance of the features without spatial pyramids 
  (CNN-B-hidden7-none in Table~\ref{tab:med14_single_feat}).
  Thus the spatial pyramid is effective in capturing useful information;
  and the feature dimension can be reduced without loss of important information.
\begin{table}[tb]
\setlength{\tabcolsep}{3 pt}
\small
\begin{center}
\begin{tabular}{|l|r|c|c|c|c|c|}
\hline
  \multirow{2}{*}{\textbf{Layer}} 
& \multirow{2}{*}{\textbf{Dim.~}}
& \multirow{2}{*}{\textbf{SP}}
& \multirow{2}{*}{\textbf{Norm}}
& \multirow{2}{*}{\textbf{SVM}}
& \multicolumn{2}{c|}{\textbf{mAP}}\\
& 
& 
& 
& 
& \textbf{CNN A}  & \textbf{CNN B}\\
\hline
output & 1,000 & none&  root & linear  & 15.90\% & 19.46\% \\ 
output & 8,000 & SP8 &  root & linear  & 22.04\% & 25.67\% \\ 
output & 8,000 & SP8 &$\ell_2$& linear & 22.01\% & 25.88\% \\ 
output & 1,000 & none&  root & $\chi^2$& 21.54\% & 25.30\% \\ 
output & 8,000 & SP8 &  root & $\chi^2$& 27.22\% & 31.24\% \\
output & 8,000 & SP8 &  root & RBF     & 27.12\% & 31.73\% \\
\hline
hidden6 & 4,096  & none& $\ell_2$ & linear & 22.11\% & 25.37\% \\ 
hidden6 & 32,768 & SP8 & root     & linear & 21.13\% & 26.27\% \\ 
hidden6 & 32,768 & SP8 & $\ell_2$ & linear & 23.21\% & 28.31\% \\ 
hidden6 & 4,096  & none& $\ell_2$ & RBF    & 28.02\% & 32.93\% \\
hidden6 & 32,768 & SP8 & $\ell_2$ & RBF    & 28.20\% & 33.54\% \\
\hline
hidden7 & 4,096  & none& $\ell_2$ & linear & 21.45\% & 25.08\% \\ 
hidden7 & 32,768 & SP8 & root     & linear & 23.72\% & 26.57\% \\ 
hidden7 & 32,768 & SP8 & $\ell_2$ & linear & 25.01\% & 29.70\%  \\ 
hidden7 & 4,096  & none& $\ell_2$ & RBF    & 27.53\% & 33.72\%  \\
hidden7 & 32,768 & SP8 & $\ell_2$ & RBF    & 29.41\% & \textbf{34.95\%}  \\
\hline
\end{tabular}
\end{center}
\caption{Various configurations of video classification.
Spatiotemporal pooling: max pooling.
A and B: 8- and 19-layer CNNs from \cite{krizhevsky2012} and~\cite{oxford_vdeep}, resp. 
SP8: the $1\times1+3\times1+2\times2$ and $3\times1$ spatial pyramids.
Dataset: TRECVID MED'14 100Ex.
}
\label{tab:med14_single_feat}
\end{table}

\begin{table}[tb]
\setlength{\tabcolsep}{5 pt}
\small
\begin{center}
\begin{tabular}{|l|c|c|c|c|}
\hline
  \textbf{Layer}
& \textbf{Dim.}
& \textbf{Norm}
& \textbf{SVM}
& \textbf{mAP}\\
\hline
output & 2,000 & root & linear  & 23.62\%  \\
output & 2,000 & root & $\chi^2$& 29.26\% \\
\hline
hidden6 & 8,192 & $\ell_2$ & linear & 26.69\% \\
hidden6 & 8,192 & $\ell_2$ & RBF    & 33.33\% \\ 
\hline
hidden7 & 8,192 & $\ell_2$ & linear & 27.41\% \\
hidden7 & 8,192 & $\ell_2$ & RBF    & 34.51\% \\ 
\hline
\end{tabular}
\end{center}
\caption{Objectness-guided pooling; CNN B; MED'14 100Ex.}
\label{tab:med14_obj}
\end{table}

\begin{table}[tb]
\setlength{\tabcolsep}{5 pt}
\small
\begin{center}
\begin{tabular}{|r|c|c|}
\hline
  \multirow{2}{*}{\textbf{Dim.~}}
& \multicolumn{2}{c|}{\textbf{SVM/mAP}} \\
& \textbf{linear}
& \textbf{RBF} \\
\hline
32,768 & 29.70\% & 34.95\% \\
\hline
4096   & 29.69\% & 34.55\% \\
2048   & 29.02\% & 34.00\% \\ 
1024   & 28.34\% & 33.18\% \\
\hline
\end{tabular}
\end{center}
\caption{Applying PCA on the top-perfoming feature
in Table~\ref{tab:med14_single_feat} prior to SVM.
Feature: hidden7-layer feature extracted from CNN architecture B with SP8 and $\ell_2$ normalization.
}
\label{tab:pca}
\end{table}

\subsection{Fusion Performance}
We evaluated fusion of CNN features and Fisher vectors (FVs). 
The mAPs (and features dimensions) obtained with D-SIFT+FV and IDT+FV using RBF SVM classifiers were $24.84\%$ and $28.45\%$ ($98,304$ and $101,376$), respectively.
Table~\ref{tab:fusion} reports our various fusion experiments. 
As expected, the late fusion of the static (D-SIFT) and motion-based (IDT) FVs brings a substantial improvement over the results obtained by the motion-based only FV. 
The fusion of CNN features does not provide much gain.
This can be explained by the similarity of the information captured by the hidden layers and the output layer. 
Similarly, fusion of the hidden layer with the static FV does not provide improvement, although the output layer can still benefit from the late fusion with the static FV feature.
However, fusion between any of the CNN-based features and the motion-based FV brings a consistent gain over the single best performer. 
This indicates that appropriate integration of motion information into the CNN architecture leads to substantial improvements. 
\begin{table}[h!]
\setlength{\tabcolsep}{5pt}
\small
\begin{center}
\begin{tabular}{|l|c|c|}
\hline
\textbf{Features} & \textbf{mAP} & \textbf{vs. single feat.} \\
\hline
D-SIFT+FV, IDT+FV             & 33.09\%           & +8.25\%, +4.64\%\\
\hline
CNN-output, CNN-hidden6       & 35.04\%         & +3.80\%, +1.50\%\\
CNN-output, CNN-hidden7       & 34.92\%         & +3.68\%, -0.03\%\\
CNN-hidden6, CNN-hidden7      & 34.85\%         & +1.31\%, -0.10\%\\
\hline
CNN-output, D-SIFT+FV         & 31.45\%         & +0.21\%, +6.61\%\\
CNN-hidden6, D-SIFT+FV        & 31.71\%         & -1.83\%, +6.87\%\\
CNN-hidden7, D-SIFT+FV        & 33.21\%         & -1.74\%, +8.37\%\\
\hline
CNN-output, IDT+FV            & 37.97\%         & +6.73\%, +9.52\%\\
CNN-hidden6, IDT+FV           & 38.30\%         & +4.76\%, +9.85\%\\
CNN-hidden7, IDT+FV           & \textbf{38.74}\%& +3.79\%, +10.29\%\\
\hline
\end{tabular}
\end{center}
\caption{Fusion. 
Used the best performing CNN features with kernel SVMs in Table~\ref{tab:med14_single_feat}.
}
\label{tab:fusion}
\end{table}

\subsection{Comparison with the State-of-the-Art}
Table~\ref{tab:med14_comp} shows the comparison of the proposed approach with strong hand-engineered approaches and other CNN-based approach on TRECVID MED'14.
The proposed approach based on CNN features significantly outperformed
strong hand-engineered approaches (D-SIFT+FV, IDT+FV and MIFS)
even without integrating motion information.
The CNN-based features have a lower dimensionality than the Fisher vectors. 
In particular, the dimension of the output layer is an order of magnitude smaller.
The CNN architecture has been trained on high resolution images, whereas we applied it on low-resolution video frames which suffer from compression artifacts and motion blur,
This further confirms that the CNN features are very robust,
despite the domain mismatch.
The proposed approach also outperforms the CNN-based approach by Xu \textit{et al.}~\cite{DBLP:journals/corr/XuYH14}.
Developed independently,
they used the same CNN architecture as ours,
but different CNN layers, pooling, feature encoding and fusion strategies.
The proposed approach outperforms all these competitive approaches and yields the new state-of-the-art,
thanks to the carefully designed system based on CNNs and the fusion with motion-based features.
\begin{table}[tb]
\setlength{\tabcolsep}{5pt}
\small
\begin{center}
\begin{tabular}{|l|c|l|}
\hline
\textbf{Method}
& \textbf{CNN}
& \textbf{mAP}\\
\hline
D-SIFT~\cite{lowe2004}+FV~\cite{perronnin2010}      & no  & 24.84\%  \\
IDT~\cite{wang2013action}+FV~\cite{perronnin2010}   & no  & 28.45\%  \\
D-SIFT+FV, IDT+FV (fusion)                          & no  & 33.09\%  \\
MIFS \cite{MIFS}                                    & no  & 29.0 \%  \\
CNN-LCD$_\text{VLAD}$ with multi-layer fusion           \cite{DBLP:journals/corr/XuYH14} & yes & 36.8 \%          \\
\hline
proposed: CNN-hidden6 & yes        & 33.54\% \\
proposed: CNN-hidden7 & yes        & 34.95\%\\
proposed: CNN-hidden7, IDT+FV & yes &\textbf{38.74}\% \\
\hline
\end{tabular}
\end{center}
\caption{Comparison with other approaches
on TRECVID MED'14 100Ex}
\label{tab:med14_comp}
\end{table}

\section{Evaluation in Action Recognition}
\subsection{Single Feature Performance}
We evaluated our approach on a well-established 
action recognition dataset UCF-101~\cite{soomro2012}.
We followed the three splits in the experiments and report overall accuracy in each split 
and the mean accuracy across three splits.
The configuration was the same as the one used in TRECVID MED'14 
experiments,
except that only linear SVM was used for fare comparison with other approaches on this dataset.
The results are given in Table~\ref{tab:ucf101_single_feat}.
The CNN architectures from \cite{krizhevsky2012} and
\cite{oxford_vdeep} are referred as A and B.
It shows that using hidden layer features yields 
better recognition accuracy compared to using softmax activations at the output layer.
The performance also improves by up to $8\%$ using deeper CNN architecture.
\begin{table}[tb]
\setlength{\tabcolsep}{5 pt}
\small
\begin{center}
\begin{tabular}{|c|l|ccc|c|}
\hline
\textbf{\makecell{CNN\\archi.}}
& \textbf{Features} 
& \textbf{\makecell{split 1\\acc.}}
& \textbf{\makecell{split 2\\acc.}}
& \textbf{\makecell{split 3\\acc.}}
& \textbf{\makecell{mean\\acc.}} \\
\hline
A&output  & 69.02\% & 68.21\% & 68.45\% & 68.56\% \\
A&hidden6 & 70.76\% & 71.21\% & 72.05\% & 71.34\% \\
A&hidden7 & 72.35\% & 71.64\% & 73.19\% & 72.39\% \\ 
\hline
B&output    & 75.65\% & 75.74\% & 76.00\% & 75.80\% \\
B&hidden6   & 79.88\% & 79.14\% & 79.00\% & 79.34\% \\
B&hidden7   & 79.01\% & 79.30\% & 78.73\% & 79.01\% \\
\hline
\end{tabular}
\end{center}
\caption{Accuracy of Single Features using proposed approach on UCF-101.
Configurations are the same as in experiments on TRECVID MED'14. 
Features are extracted with SP8.
All features are classified with linear SVMs.
}
\label{tab:ucf101_single_feat}
\end{table}

\subsection{Fusion Performance}
We extracted IDT+FV features on this dataset and obtained mean accuracy of $86.5\%$ across three splits.
Unlike the event detection task in TRECVID MED'14 dataset,
the action recognition in UCF-101 is temporally trimmed and more centered on motion.
Thus, the motion-based IDT+FV approach outperformed the image-based CNN-approach.
However, as shown in Table~\ref{tab:ucf101_fusion}, a simple weighted average fusion of CNN-hidden6 and IDT+FT features boosts the performance to the best accuracy of $89.62\%$.
\begin{table}[tb]
\setlength{\tabcolsep}{5pt}
\small
\begin{center}
\begin{tabular}{|c|l|ccc|c|}
\hline
\textbf{\makecell{CNN\\archi.}}
& \textbf{\makecell{CNN\\layer}}
& \textbf{\makecell{split 1\\acc.}}
& \textbf{\makecell{split 2\\acc.}}
& \textbf{\makecell{split 3\\acc.}}
& \textbf{\makecell{mean\\acc.}} \\
\hline
A&output
& 85.57\% & 87.01\% & 87.18\% & 86.59\% \\
A&hidden6
& 86.39\% & 87.36\% & 87.61\% & 87.12\% \\
A&hidden7
& 86.15\% & 87.98\% & 87.74\% & 87.29\% \\
\hline
B&output
& 87.95\% & 88.43\% & 89.37\% & 88.58\% \\
B&hidden6
& 88.63\% & 90.01\% & 90.21\% & \textbf{89.62}\% \\
B&hidden7
& 88.50\% & 89.29\% & 90.10\% & 89.30\% \\
\hline
\end{tabular}
\end{center}
\caption{Accuracy of late fusion of 
CNN features and IDT+FV features. 
The configurations are the same as in Table~\ref{tab:ucf101_single_feat}.
We extracted IDT+FV features based on \cite{wang2013action,perronnin2010}.
}
\label{tab:ucf101_fusion}
\end{table}

\subsection{Comparison with the State-of-the-Art}
Table~\ref{tab:ucf101_comp} shows comparisons with other approaches based on neural networks.
Note that the proposed image-based CNN-approach yields superior performance (6.3\%, 6.0\% and 13.9\% higher) than 
the spatial stream ConvNet in \cite{karen-flow},
the single-frame model in \cite{Ng2015LSTM}
and the slow-fusion spatiotemporal ConvNet in \cite{karpathy2014}
even though our model is not fine-tuned on specific dataset.
The fusion performance of CNN-hidden6 (or CNN-hidden7) and IDT+FV features 
also outperforms 
the two stream CNN approach \cite{karen-flow}  
and long short term memory (LSTM) approach
that utilizing both image and optical flow \cite{Ng2015LSTM}.
\begin{table}[tb]
\setlength{\tabcolsep}{5pt}
\small
\begin{center}
\begin{tabular}{|lc|}
\hline
\textbf{Method}
\textbf{acc.} \\
\hline
Spatial stream ConvNet~\cite{karen-flow}            & 73.0\%  \\
Temporal stream ConvNet~\cite{karen-flow}           & 83.7\%  \\
Two-stream ConvNet fusion by avg~\cite{karen-flow}  & 86.9\%  \\  
Two-stream ConvNet fusion by SVM~\cite{karen-flow}  & 88.0\%  \\  
Slow-fusion spatiotemporal ConvNet \cite{karpathy2014}& 65.4\%  \\  
Single-frame model \cite{Ng2015LSTM}                & 73.3\%  \\  
LSTM (image + optical flow) \cite{Ng2015LSTM}       & 88.6\%  \\  
\hline
proposed: CNN-hidden6 only                          & 79.34\% \\
proposed: CNN-hidden6, IDT+FV (avg. fusion)         & \textbf{89.62}\% \\
proposed: CNN-hidden7, IDT+FV (avg. fusion)         & 89.30\% \\
\hline
\end{tabular}
\end{center}
\caption{Comparison with other approaches based on neural networks in mean accuracy over three splits on UCF-101}
\label{tab:ucf101_comp}
\end{table}

\section{Computational Cost}
\label{sec:computationalCost}
We have benchmarked the extraction time of the Fisher vectors and CNN features on a CPU machine. Extracting D-SIFT (resp. IDT) Fisher vector takes about $0.4$ (resp. $5$) times the video playback time, while the extraction of the 
CNN features requires $0.4$ times the video playback time. 
The CNN features can thus be extracted in real time. 
On the classifier training side, it requires about 150s to train a kernel SVM event detector using the Fisher vectors, while it takes around 90s with the CNN features using the same training pipeline. 
On the testing side, it requires around 30s to apply a Fisher vector trained event model on the 23,953 TRECVID MED'14 videos, while it takes about 15s to apply a CNN trained event model on the same set of videos.    

\section{Conclusion}
\label{sec:conclusion}
In this paper we proposed a step-by-step procedure to fully exploit the potential of image-trained CNN architectures for video classification. While every step of our procedure has an impact on the final classification performance, we showed that CNN architecture, the choice of CNN layer, the spatiotemporal pooling, the normalization, and the choice of classifier are the most sensitive factors. Using the proposed procedure, we showed that an image-trained CNN architecture can outperform 
competitive \emph{motion-} and \emph{spatiotemporal-} based non-CNN approaches
on the challenging TRECVID MED'14 video dataset. 
The result shows that improvements on the image-trained CNN architecture are also beneficial to video classification, despite the domain mismatch. 
Moreover, we demonstrated that adding some motion-information via late fusion brings substantial gains,
outperforming other vision-based approaches on this MED'14 dataset.
Finally, the proposed approach is compared with other neural network approaches on the action recognition dataset UCF-101.
The image-trained CNN approach is comparable with the state-of-the-art
and the late fusion of image-trained CNN features and motion-based IDT-FV features outperforms the state-of-the-art.

In this work we used an image-trained CNN as a black-box feature extractor.
Therefore, we expect any improvements in the CNN to directly lead to improvements in video
classification as well. The CNN was trained on the ImageNet dataset which mostly
contains high resolution photographic images whereas the video dataset is fairly
heterogeneous in terms of quality, resolution, compression artifacts and camera
motion. Due to this domain mismatch, we believe that additional gains
can be achieved by fine-tuning the CNN for the dataset. Even more improvements
can possibly be made by learning motion information through a spatiotemporal deep neural network 
architecture.


{\small
\bibliographystyle{ieee}
\bibliography{bib/papers.bib,bib/source_code.bib}
}
\end{document}